\documentclass[11pt]{article}
\usepackage[final]{acl}
\usepackage{times}
\usepackage{latexsym}
\usepackage[T1]{fontenc}
\usepackage[utf8]{inputenc}
\usepackage{microtype}
\usepackage{inconsolata}
\usepackage{graphicx}
\usepackage{todonotes}
\usepackage{comment}
\usepackage{booktabs}
\usepackage{multirow}
\usepackage{subcaption} 
\usepackage{amssymb}
\usepackage{amsmath}
\usepackage{mathtools}

\usepackage[most]{tcolorbox}
\usepackage{xcolor}
\usepackage{fancyvrb}
\usepackage{fvextra}

\definecolor{promptbg}{RGB}{248,249,252}
\definecolor{promptframe}{RGB}{70,90,140}

\newtcolorbox{promptbox}[1]{
  enhanced, breakable,
  colback=promptbg,
  colframe=promptframe,
  coltitle=white,
  fonttitle=\bfseries\small,
  title=#1,
  boxrule=0.4pt,
  left=4pt, right=4pt, top=3pt, bottom=3pt,
  arc=2pt,
}

\title{Grounding Memory Summarization in Utility Intent}

\author{
    Zhenyu Lei$^{\blacklozenge}$ \:
    Mingjia Shi$^\blacklozenge$ \:
    Xingbo Fu$^\blacklozenge$ \\ \bf
    Haoyu He$^\spadesuit$ \:
    Qi R. Wang$^\spadesuit$ \:
    Jundong Li$^\blacklozenge$ \\
    $^\blacklozenge$University of Virginia, $^\spadesuit$Northeastern University\\
    \texttt{\{vjd5zr, nzh3ru, xf3av, jundong\}@virginia.edu}\\
    \texttt{\{he.haoyu1, q.wang\}@northeastern.edu}
}

\begin{document}
\maketitle

\begin{abstract}
Existing summarizers for memory systems are typically optimized for human-facing criteria such as faithfulness, which misaligns with their true objective: preserving the evidence needed to support future queries. We show that conditioning summarization on query-answer pairs substantially improves answer quality, and that this utility-aware behavior is transferable across queries. Motivated by these findings, we propose MemSuit, a self-distillation framework in which a teacher summarizer, conditioned on observed query-answer pairs, produces utility-aware memory entries that a student learns to reproduce from the raw conversation alone. To prevent collateral erasure where conditioning on a single query-answer pair discards evidence relevant to other plausible queries, the teacher decomposes each block into multiple self-contained entries that preserve distinct query-relevant facets as independently retrievable units. To align the retriever with the compact, fact-dense style of teacher entries, we further fine-tune the embedding model with a contrastive objective supervised by teacher entries. Across a diverse suite of conversational query types, MemSuit consistently outperforms state-of-the-art baselines, confirming the value of grounding memory in downstream utility. 
\end{abstract}

\begin{figure}[t]
    \centering
    
    \begin{subfigure}{\linewidth}
        \centering
        \includegraphics[width=\linewidth]{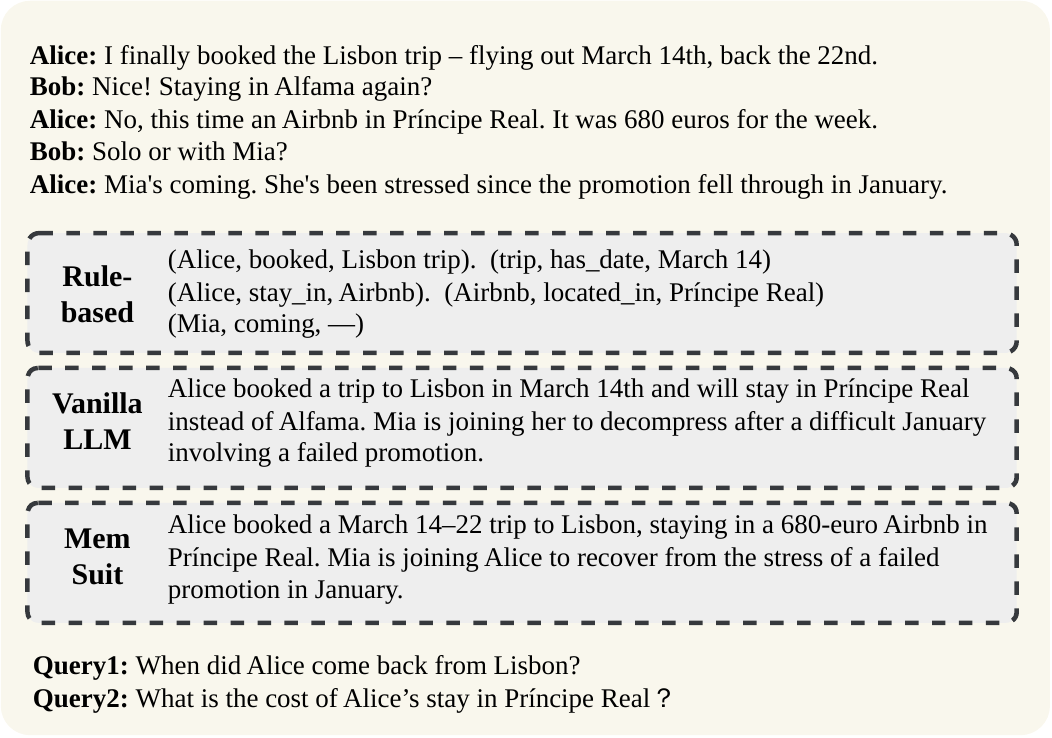}
        \caption{Summary Case Study}
        \label{fig:prelim_study}
    \end{subfigure}

    \begin{subfigure}{\linewidth}
        \centering
        \includegraphics[width=\linewidth]{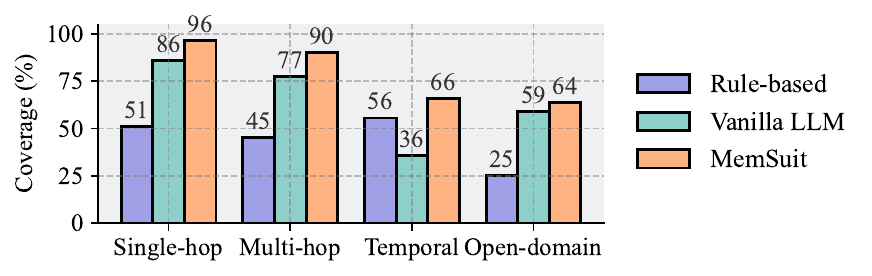}
        \caption{Evidence Coverage}
        \label{fig:vanilla_vs_utility}
    \end{subfigure}

    \caption{
    (a) A short conversation summarized by different methods. Neither the rule-based nor the vanilla LLM summary retains enough details to answer the two example queries.
    (b) Evidence coverage on LoCoMo (Qwen2.5-3B), measured as the fraction of query-supporting facts preserved in the summary. Vanilla LLM summaries lose substantial evidence for temporal and open-domain queries.}
    \label{fig:teaser}
\end{figure}

\section{Introduction}
Memory systems have become a standard component of LLM agents deployed in long-horizon interactive settings, storing past interactions and retrieving relevant fragments on demand~\cite{packer2023memgpt,edge2024local, zhang2025survey}. Since storing raw conversational text verbatim is wasteful where the vast majority of utterances are rarely relevant for future queries~\cite{maharana2024evaluating}, a substantial body of work has focused on memory summarization: compressing raw history into compact yet informative representations~\cite{wang2025recursively, hu2025hiagent}. Existing approaches fall into two broad families. Extraction-based methods retain predefined patterns such as entities, events, or temporal markers through handcrafted heuristics, but they struggle to adapt across diverse conversational styles~\cite{liu2025integrating, gupta2019abstractive}. LLM-based methods, which have become the de facto solution, instead prompt a language model to retain content it deems important~\cite{zhong2024memorybank,chhikara2025mem0}. However, as illustrated in Figure~\ref{fig:teaser}, these methods are prone to preserving topic-level gist while discarding the concrete details on which downstream demand most often hinges. As a result, the evidence preserved in memory is often insufficient for downstream answering, particularly for temporal and open-domain queries that depend on fine-grained and implicit information from past interactions.

While prior work has sought to improve memory summarization with various prompts~\cite{xu2026mem, liu2026simplemem}, we argue that this shortcoming reflects a more fundamental issue: a misalignment of objectives. LLM summarizers are typically trained against human-written reference summaries, which optimize for properties such as coverage, coherence, and topical faithfulness~\cite{zhang2024benchmarking}. However, in a memory pipeline, a summary is consumed by a retrieval-and-answer system, and its value is more determined by whether it preserves the information needed to answer future queries, a criterion of downstream utility rather than human-facing fidelity.
This diagnosis is corroborated by a simple oracle analysis: when summarizers are granted privileged access to future query-answer pairs and instructed to retain only the evidence needed to support each answer, the resulting memory entries exhibit substantially higher downstream performance. However, such an oracle setting illustrates only the ideal upper bound and is unrealistic in practice, where summaries must be generated and committed to storage hundreds or thousands of turns before future queries are observed~\cite{yang2026beyond}.

In this paper, we demonstrate that utility-aware summarization patterns are transferable across conversations and queries: the criteria a summarizer learns to apply when the query is known generalize to settings where the query is hidden. Motivated by this insight, we propose a downstream-utility-oriented memory summarization framework based on self-distillation. Our approach instantiates two summarization policies within a single pipeline. A teacher model is granted access to observed query-answer pairs and generates tailored summaries, retaining precisely the information required to address the specific question at hand. Conversely, a student model only observes the raw conversation, mirroring the deployment setting where future queries are unknown, and is trained to imitate the teacher across the observed query distribution. Intuitively, the student learns to anticipate what information would be relevant for plausible future queries, thereby internalizing the teacher's utility-aware behavior into a single, query-agnostic summarization policy.

However, operationalizing this principle surfaces two obstacles that a naive student-teacher pipeline cannot overcome. The first we term \textbf{collateral erasure}: a single conversational block rarely revolves around a single topic, and the topics it threads together may each anchor details for different queries~\cite{zou2026mem}. Compressing such a block into one utility-aware summary inevitably privileges the conditioning query-answer pair, causing the details that would have supported other plausible queries to be discarded as incidental. The second is \textbf{retrieval mismatch}: teacher-produced summaries are deliberately terse and fact-dense, whereas off-the-shelf embedding models for retrieval are calibrated on the verbose, redundant prose of naturally occurring text~\cite{li2025survey}. Even when the summaries contain the right information, this gap prevents the retriever from finding them, capping end-to-end performance.

We address both obstacles within a unified framework MemSuit (\textbf{Mem}ory \textbf{Su}mmarization for Util\textbf{it}y). To prevent collateral erasure, MemSuit allows the teacher to emit a variable number of memory entries per block: rather than compressing a pre-defined block into a single summary, the teacher segments it along semantic boundaries and writes one self-contained entry per topic. Distinct query-relevant aspects of the conversation are thereby preserved in independently retrievable entries, mitigating the interference induced by conditioning on a single query-answer pair. To close the retrieval mismatch, we fine-tune the embedding model with a contrastive objective: teacher-generated entries serve as positives for their conditioning queries and unrelated entries as negatives, aligning the embedding space with our compact summaries without requiring additional annotations. Across a diverse suite of conversational query types, MemSuit consistently outperforms state-of-the-art baselines, confirming the value of grounding memory in downstream utility.

\section{Preliminaries}
\label{sec:preliminary}

\subsection{Problem Formulation}
\label{subsec:formulation}

We consider long-horizon conversational memory. Let $\mathcal{C} = (u_1, u_2, \ldots, u_T)$ denote a conversation of $T$ utterances, and let $\mathcal{Q} = \{(q_i, a_i)\}_{i=1}^{N}$ denote a set of query--answer pairs grounded in $\mathcal{C}$. A memory system comprises three components: a \emph{summarizer} $f_\theta$ that compresses $\mathcal{C}$ into a set of $N$ memory entries $\mathcal{M} = f_\theta(\mathcal{C}) = \{m_1, \ldots, m_N\}$; a \emph{retriever} $g_\phi$ that, given a query $q$, returns a relevant subset $\mathcal{M}_q \subseteq \mathcal{M}$; and a fixed \emph{reader} $\pi$ that produces an answer $\hat{a} = \pi(q, \mathcal{M}_q)$. The goal is to design and learn the memory system such that the expected answer quality $\mathbb{E}_{(q,a) \sim \mathcal{Q}}\bigl[\mathrm{score}(\hat{a}, a)\bigr]$ is maximized, where $\mathrm{score}(\cdot,\cdot)$ denotes a task-specific metric.

\subsection{Preliminary Study}
\label{subsec:prelim_study}

Our approach rests on two empirical claims: (i) conditioning summarization on downstream utility yields memory entries materially more useful for question answering than query-agnostic summaries, and (ii) the underlying utility-aware behavior is transferable, allowing a summarizer to internalize it without access to the target query at storage time. We examine each claim through a simple study on the LoCoMo benchmark~\cite{maharana2024evaluating}, using \texttt{Qwen2.5-3B-Instruct} as both the summarizer and the downstream reader. Following common practice in conversational memory systems, the summarizer is applied at the granularity of \emph{blocks}: contiguous spans $b \subseteq \mathcal{C}$ of consecutive utterances. All results are F1 scores averaged over 500 randomly sampled queries.

\paragraph{Study 1: Does utility-awareness improve summary quality?}
We first ask whether exposing the summarizer to a downstream query-answer pair improves the memory entries it produces. For each sampled query $q$ with reference answer $a$, we locate the contiguous block $b \subseteq \mathcal{C}$ that grounds $(q, a)$, summarize $b$ under one of two conditions, and have the reader answer $q$ from the resulting entry. Under the \emph{vanilla} condition, the summarizer receives only $b$, mirroring standard practice in prior memory systems. Under the \emph{utility-aware} condition, it additionally observes $(q, a)$ and is instructed to retain the evidence only from $b$ required to support $a$.

As shown in Figure~\ref{fig:vanilla_vs_utility}, utility-aware summaries yield substantial F1 gains across all types of queries, nearly doubling overall performance. This indicates that one of the bottleneck of standard summarization is the objective alignment: once the summarizer knows what the summary is for, it preserves the right evidence. However, this condition is not itself deployable, since it presupposes access to $(q, a)$ at storage time.

\begin{figure}[t]
    \centering
    
    \begin{subfigure}{\linewidth}
        \centering
        \includegraphics[width=\linewidth]{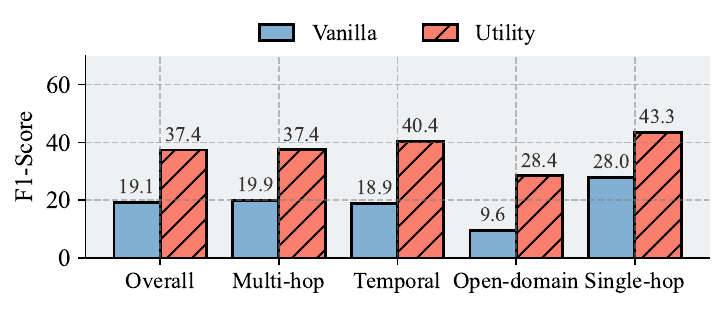}
        \caption{Vanilla vs. utility-aware summarization}
        \label{fig:vanilla_vs_utility}
    \end{subfigure}
    
    \begin{subfigure}{\linewidth}
        \centering
        \includegraphics[width=\linewidth]{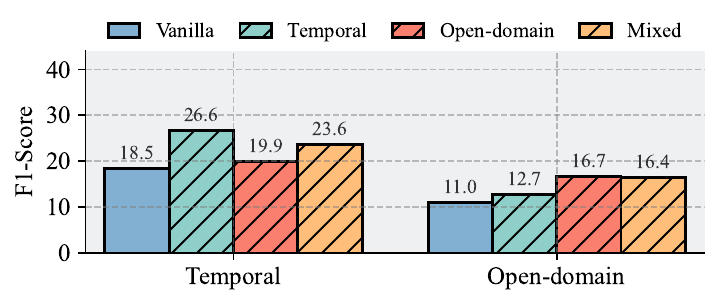}
        \caption{Transferability across demonstration types}
        \label{fig:prelim_study}
    \end{subfigure}
    \caption{
    (a) Granting the summarizer access to the downstream query-answer pair (\textit{Utility}) substantially outperforms the query-agnostic baseline (\textit{Vanilla}). 
    (b) In-context utility-aware demonstrations surpasses the vanilla baseline. Gains peak when demonstration and evaluation types match and persist under mixing, indicating that utility-aware patterns transfer across queries}
    \vspace{-10pt}
    \label{fig:pre}
\end{figure}

\paragraph{Study 2: Does utility-aware behavior transfer across queries?}
Study 1 establishes an unrealizable upper bound. As a result, we next ask whether the criteria implicit in utility-aware summarization can transfer to a query-agnostic setting through demonstration alone. For each sampled query $q$, we summarize its grounding block $b$ under a demo-conditioned prompt that hides $q$ but provides in-context demonstrations, each a quadruple $(q', a', b', m'_{\text{util}})$ comprising a different query, its answer, the grounding block, and the corresponding utility-aware summary; the reader then answers $q$ from the resulting summary. We vary two factors. The demonstration type (Temporal, Open-domain, or a Mix) controls the kind of utility-aware patterns the model observes, and the evaluation type specifies the category of $q$ at test time.

Figure~\ref{fig:prelim_study} reveals two findings. First, every demo-conditioned setting outperforms the vanilla baseline, showing that utility-aware pattern is transferable, where the model generalizes the criteria implicit in the demonstrations to unseen queries. Second, transfer is stronger when demonstration and evaluation types match, while mixing in off-category demonstrations largely preserves these in-category gains rather than washing them out. While in-context demonstrations are able to elicit some transfer, this approach is fragile in deployment as it depends on demonstration selection and consumes prompt budget at every storage step. We instead pursue a parametric solution with self-distillation to internalize the utility-aware pattern.

\begin{figure*}
    \centering
    \includegraphics[width=\linewidth]{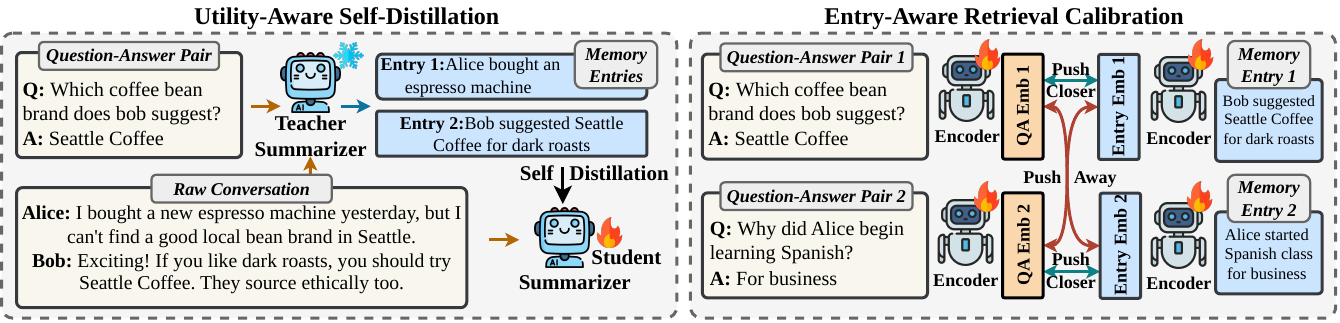}
    \caption{Overview of MemSuit. \textit{Left:} A teacher summarizer conditioned on observed Q-A pairs decomposes each conversation into utility-aware memory entries; a student is distilled to reproduce these entries from the raw conversation alone. \textit{Right:} A contrastive objective calibrates the retriever's embedding space, pulling each query toward its grounding entries and pushing it away from entries grounding other queries.}
    \label{fig:overview}
\end{figure*}

\section{Method}
\label{sec:method}

Guided by the findings in Section~\ref{sec:preliminary}, \textbf{MemSuit} operationalizes utility-aware summarization through self-distillation. A utility-aware \emph{teacher}, conditioned on observed query--answer pairs, produces memory entries that retain precisely the evidence required to support each answer. A query-agnostic \emph{student} then learns to imitate the teacher from the raw conversation alone, internalizing its selection criteria into a single deployable summarization policy.
MemSuit instantiates this scheme in three stages. We first introduce \textbf{Adaptive Entry Decomposition} (Section~\ref{subsec:entry-split}), which lets the teacher emit a variable-length sequence of self-contained entries per block, addressing \emph{collateral erasure} at the source of supervision. We then describe \textbf{Utility-Aware Self-Distillation} (Section~\ref{subsec:distillation}), which trains the student to reproduce these decomposed entries from the block alone. Finally, \textbf{Entry-Aware Retrieval Calibration} (Section~\ref{subsec:contrastive}) closes the \emph{retrieval mismatch} by reusing the teacher's entries as contrastive supervision for the embedding model. Figure~\ref{fig:overview} provides an overview.

\subsection{Adaptive Entry Decomposition}
\label{subsec:entry-split}

We begin by specifying how the teacher constructs memory entries, since these entries serve as the supervisory target for the student summarizer.
Let $f^{\mathrm{T}}$ denote the teacher summarizer, which is invoked only at training time and conditions on observed query--answer pairs. For a conversation block $b \subseteq \mathcal{C}$, let
$\mathcal{Q}_b \coloneqq \{(q,a) \in \mathcal{Q} \mid (q,a) \text{ is grounded in } b\}$
denote the set of query--answer pairs grounded in $b$. Given any pair $(q,a) \in \mathcal{Q}_b$, the teacher produces utility-aware content that retains the evidence in $b$ required to support $a$.
A naive instantiation would compress $b$ into a single summary $m = f^{\mathrm{T}}(b, q, a)$. This invites \emph{collateral erasure}: conditioning on $(q,a)$ biases the summary toward evidence for this specific pair, so content supporting other plausible pairs $(q', a') \in \mathcal{Q}_b \setminus \{(q,a)\}$ is discarded as incidental.
To mitigate this, we allow the teacher to jointly segment $b$ and produce a variable-length set of self-contained entries in a single pass:
\begin{equation}
\resizebox{\linewidth}{!}{$
    \mathcal{M}^{\mathrm{T}}_{b,q} = \bigl\{m^{\mathrm{T}}_{1}, \ldots, m^{\mathrm{T}}_{n_{b,q}}\bigr\} = f^{\mathrm{T}}(b, q, a), \quad n_{b,q} \geq 1.
$}
\end{equation}
where each $m^{\mathrm{T}}_{j}$ summarizes a distinct semantically coherent region of $b$. Within this set, at least one entry $m^{\mathrm{T}}_{j^\star}$ is anchored to the observed pair $(q, a)$ and retains the evidence required to support $a$, while the remaining entries $\{m^{\mathrm{T}}_{j}\}_{j \neq j^\star}$ summarize orthogonal regions of $b$ without query conditioning.
This decomposition preserves utility-aware evidence for the observed query at $m^{\mathrm{T}}_{j^\star}$ while shielding the rest of $b$ from query-induced bias, ensuring that information relevant to other potential pairs in $\mathcal{Q}_b$ remains independently retrievable downstream. The prompt used to elicit this joint decomposition is provided in the Appendix.

\subsection{Utility-Aware Self-Distillation}
\label{subsec:distillation}

With the teacher's decomposed entries in hand, we distill its utility-aware behavior into a student summarizer $f^{\mathrm{S}}_{\theta}$ that, unlike the teacher, is conditioned solely on the block $b$ and never observes $(q, a)$.
For each training block $b$ and grounded pair $(q, a) \in \mathcal{Q}_b$, we treat the teacher's decomposed entries $\mathcal{M}^{\mathrm{T}}_{b,q}$ as the supervisory target. Concretely, we serialize them into a single delimited sequence,
\begin{equation}
    y_{b,q} \;=\; m^{\mathrm{T}}_{1} \,\Vert\, m^{\mathrm{T}}_{2} \,\Vert\, \cdots \,\Vert\, m^{\mathrm{T}}_{n_{b,q}},
\end{equation}
where $\Vert$ denotes concatenation with a special delimiter token, and optimize the student with a token-level cross-entropy objective:
\begin{equation}
\resizebox{\linewidth}{!}{$
    \mathcal{L}_{\mathrm{distill}}(\theta) = -\,\mathbb{E}_{(b,q) \sim \mathcal{D}_{\mathrm{tr}}} \!\left[ \sum_{t=1}^{|y_{b,q}|} \log p_{\theta}\!\left(y_{b,q}^{(t)} \,\middle|\, y_{b,q}^{(<t)},\, b \right) \right].
$}
\end{equation}
where $\mathcal{D}_{\mathrm{tr}} = \{(b, q, a) : b \subseteq \mathcal{C},\, (q,a) \in \mathcal{Q}_b\}$ denotes the training distribution and $y_{b,q}^{(t)}$ is the $t$-th token of $y_{b,q}$.
Crucially, by structuring the target as a delimited sequence of self-contained entries rather than a monolithic summary, the student learns not only \emph{what} content the teacher retains for plausible future queries, but also \emph{how} the teacher partitions a block so that distinct query-relevant facets remain independently retrievable.

\subsection{Entry-Aware Retrieval Calibration}
\label{subsec:contrastive}

A well-trained student is bottlenecked by the retriever: if $g_{\phi}$ cannot surface the right entry, the reader never sees it. Off-the-shelf embedding models are calibrated on verbose natural prose and tend to underperform on the compact, fact-dense entries our teacher produces. We close this gap by reusing the teacher's summaries as supervision for the retriever. Concretely, we instantiate $g_{\phi}$ as a dense retriever built on an embedding encoder $h_{\phi} : \mathcal{T} \to \mathbb{R}^{d}$ that maps any text $t \in \mathcal{T}$ (query or entry) to a $d$-dimensional vector.

\paragraph{Identifying grounding entries.}
The teacher decomposes each block $b$ into the entry set $\mathcal{M}^{\mathrm{T}}_{b,q}$, but does not explicitly mark which entries ultimately ground the answer $a$. To obtain this supervision, we prompt an LLM judge $\psi$ to identify, for each grounded pair $(q, a) \in \mathcal{Q}_b$, the subset of entries containing the evidence that supports $a$:
\begin{equation}
    \mathcal{P}_{q} \;=\; \psi\!\left(b, q, a, \mathcal{M}^{\mathrm{T}}_{b,q}\right) \;\subseteq\; \mathcal{M}^{\mathrm{T}}_{b,q}.
\end{equation}
By construction, $\mathcal{P}_{q} \neq \emptyset$: the query-anchored entry $m^{\mathrm{T}}_{j^\star}$ from Section~\ref{subsec:entry-split} is guaranteed to support $a$. The set $\mathcal{P}_{q}$ thus provides per-query positives that we use to supervise the retriever.

\paragraph{Contrastive objective.}
Given these positives, we calibrate the retriever so that each query embeds close to the entries that actually support its answer. We measure query--entry affinity via cosine similarity in the embedding space,
\begin{equation}
    s_{\phi}(q, m) \;=\; \frac{\langle h_{\phi}(q),\, h_{\phi}(m) \rangle}{\lVert h_{\phi}(q) \rVert \, \lVert h_{\phi}(m) \rVert},
\end{equation}
and contrast each query against its positives and a set of negatives drawn from unrelated blocks. Concretely, for each query $q$ in a training mini-batch $\mathcal{Q}^{\mathrm{batch}}$, let $b(q)$ denote the block in which $q$ is grounded; the in-batch negatives for $q$ are then
\begin{equation}
    \mathcal{N}_{q} \;=\; \!\!\!\bigcup_{\substack{q' \in \mathcal{Q}^{\mathrm{batch}} \\ b(q') \neq b(q)}}\!\!\! \mathcal{M}^{\mathrm{T}}_{b(q'),\,q'},
\end{equation}
and the full candidate set is $\mathcal{S}_{q} = \mathcal{P}_{q} \cup \mathcal{N}_{q}$. We optimize an InfoNCE objective:
\begin{equation}
\resizebox{\linewidth}{!}{$
    \mathcal{L}_{\mathrm{ctr}}(\phi) = -\,\mathbb{E}_{q \sim \mathcal{Q}^{\mathrm{batch}}} \log \frac{\sum_{m^{+} \in \mathcal{P}_{q}} \exp\!\bigl(s_{\phi}(q, m^{+})/\tau\bigr)}{\sum_{m \in \mathcal{S}_{q}} \exp\!\bigl(s_{\phi}(q, m)/\tau\bigr)},
$}
\end{equation}
where $\tau > 0$ is a temperature hyperparameter. This objective pulls queries toward the compact entries that actually support their answers and pushes them away from  non-grounding entries, aligning the embedding space with the distributional properties of teacher-generated memory.

\subsection{Training and Inference}
\label{subsec:training}

\paragraph{Training.}
We train the summarizer and the retriever sequentially. We first run the teacher offline over $\mathcal{D}_{\mathrm{tr}}$ to obtain the entry sets $\mathcal{M}^{\mathrm{T}}_{b,q} = f^{\mathrm{T}}(b, q, a)$ and judge-assigned positives $\mathcal{P}_{q}$ for each $(b, q, a)$. The student and the retriever are then optimized against the same supervision: $\theta^{\star} = \arg\min_{\theta} \mathcal{L}_{\mathrm{distill}}(\theta)$ and $\phi^{\star} = \arg\min_{\phi} \mathcal{L}_{\mathrm{ctr}}(\phi)$.

\paragraph{Inference.}
At deployment, neither the teacher nor any query is available at storage time. For each incoming block $b$, the student emits entries $\mathcal{M}^{\mathrm{S}}_{b} = f^{\mathrm{S}}_{\theta^{\star}}(b)$, which are encoded by $h_{\phi^{\star}}$ and appended to a growing store $\mathcal{M} = \bigcup_{b} \mathcal{M}^{\mathrm{S}}_{b}$. Given a query $q$, the retriever returns the top-$K$ entries $\mathcal{M}_{q} = \operatorname{Top\text{-}}K_{m \in \mathcal{M}}\, s_{\phi^{\star}}(q, m)$, and the fixed reader produces the answer $\hat{a} = \pi(q, \mathcal{M}_{q})$.

\section{Experiments}
In this section, we aim to address five research questions:
\textbf{RQ1:} How does MemSuit compare against state-of-the-art memory systems?
\textbf{RQ2:} What is the contribution of each component to overall performance?
\textbf{RQ3:} How does Entry-Aware Retrieval Calibration improve retrieval?
\textbf{RQ4:} Does MemSuit generalize to unseen datasets with different query distributions?
\textbf{RQ5:} How token-efficient is MemSuit at storage time?

\subsection{Experimental Setup}

\begin{table*}[t]
\centering
\renewcommand{\arraystretch}{1.15}
\setlength{\tabcolsep}{4pt}
\resizebox{\textwidth}{!}{%
\begin{tabular}{ll cc cc cc cc cc}
\toprule
\multirow{2}{*}{\textbf{Model}} & \multirow{2}{*}{\textbf{Method}} & \multicolumn{2}{c}{\textbf{Multi}} & \multicolumn{2}{c}{\textbf{Temp}} & \multicolumn{2}{c}{\textbf{Open}} & \multicolumn{2}{c}{\textbf{Single}} & \multicolumn{2}{c}{\textbf{Avg}} \\
\cmidrule(lr){3-4} \cmidrule(lr){5-6} \cmidrule(lr){7-8} \cmidrule(lr){9-10} \cmidrule(lr){11-12}
 & & \textbf{F1} & \textbf{BLEU} & \textbf{F1} & \textbf{BLEU} & \textbf{F1} & \textbf{BLEU} & \textbf{F1} & \textbf{BLEU} & \textbf{F1} & \textbf{BLEU} \\
\midrule
\multirow{6}{*}{\textbf{Llama-3.1-8B-Instruct}}
 & LoCoMo      & 21.33 & 16.42 & 17.46 & 14.41 & 10.60 & 6.69 & 30.82 & 23.41 & 20.05 & 15.23 \\
 & MemoryBank  & 24.60 & 19.46 & 12.01 & 9.56  & 10.47 & 8.18 & 38.45 & 34.34 & 21.38 & 17.89 \\
 & A-Mem       & 17.08 & 13.21 & 24.00 & 18.26 & 11.93 & 10.48 & 29.62 & 21.28 & 20.66 & 15.81 \\
 & Mem0        & 28.98 & 23.35 & 19.93 & 15.18 & 19.73 & 16.41 & \textbf{45.76} & \textbf{39.79} & 28.60 & 23.68 \\
 & SimpleMem   & 26.92 & 17.66 & 35.11 & 23.58 & 22.41 & 16.89 & 35.92 & 30.83 & 30.09 & 22.24 \\
 & \textbf{MemSuit} & \textbf{31.18} & \textbf{23.55} & \textbf{40.77} & \textbf{27.94} & \textbf{25.42} & \textbf{20.83} & 44.89 & 38.61 & \textbf{35.56} & \textbf{27.73} \\
\midrule
\multirow{6}{*}{\textbf{Qwen2.5-3B-Instruct}}
 & LoCoMo      & 16.38 & 12.44 & 16.45 & 13.31 & 4.75  & 2.84  & 18.57 & 12.64 & 14.04 & 10.31 \\
 & MemoryBank  & 20.00 & 15.22 & 15.62 & 12.22 & 8.08  & 6.33  & 27.22 & 21.94 & 17.73 & 13.93 \\
 & A-Mem       & 18.18 & 10.93 & 19.79 & 15.31 & 9.60  & 9.05  & 29.51 & 25.05 & 19.27 & 15.09 \\
 & Mem0        & 22.24 & \textbf{16.58} & 23.23 & 18.85 & 10.02 & 7.97  & 29.76 & 25.67 & 21.31 & 17.27 \\
 & SimpleMem   & 18.83 & 14.59 & 16.55 & 11.77 & 12.44 & 10.47 & 26.92 & 23.23 & 18.68 & 15.02 \\
 & \textbf{MemSuit} & \textbf{22.63} & 16.36 & \textbf{28.66} & \textbf{20.40} & \textbf{15.40} & \textbf{13.15} & \textbf{33.48} & \textbf{29.96} & \textbf{25.04} & \textbf{19.97} \\
\bottomrule
\end{tabular}%
}
\caption{Main Results. MemSuit achieves the best average performance.}
\label{tab:main}
\end{table*}

\begin{table*}[t]
\centering
\renewcommand{\arraystretch}{1.15}
\setlength{\tabcolsep}{4pt}
\resizebox{\textwidth}{!}{%
\begin{tabular}{ll cc cc cc cc cc}
\toprule
\multirow{2}{*}{\textbf{Model}} & \multirow{2}{*}{\textbf{Variant}} & \multicolumn{2}{c}{\textbf{Multi}} & \multicolumn{2}{c}{\textbf{Temp}} & \multicolumn{2}{c}{\textbf{Open}} & \multicolumn{2}{c}{\textbf{Single}} & \multicolumn{2}{c}{\textbf{Avg}} \\
\cmidrule(lr){3-4} \cmidrule(lr){5-6} \cmidrule(lr){7-8} \cmidrule(lr){9-10} \cmidrule(lr){11-12}
 & & \textbf{F1} & \textbf{BLEU} & \textbf{F1} & \textbf{BLEU} & \textbf{F1} & \textbf{BLEU} & \textbf{F1} & \textbf{BLEU} & \textbf{F1} & \textbf{BLEU} \\
\midrule
\multirow{4}{*}{\textbf{Llama-3.1-8B-Instruct}}
 & w/o Rtr-Cal     & 26.02 & 18.36 & \textbf{41.03} & 28.19 & 21.16 & 17.29 & 38.86 & 33.17 & 31.77 & 24.25 \\
 & w/o Self-Dist   & 30.92 & 21.68 & 40.24 & \textbf{28.38} & 21.36 & 16.72 & 40.92 & 36.02 & 33.36 & 25.70 \\
 & w/o Cal \& Dist & 23.16 & 15.51 & 30.88 & 21.30 & 17.53 & 14.50 & 33.38 & 28.62 & 26.24 & 19.98 \\
 & \textbf{MemSuit} & \textbf{31.18} & \textbf{23.55} & 40.77 & 27.94 & \textbf{25.42} & \textbf{20.83} & \textbf{44.89} & \textbf{38.61} & \textbf{35.56} & \textbf{27.73} \\
 \midrule
\multirow{4}{*}{\textbf{Qwen2.5-3B-Instruct}}
 & w/o Rtr-Cal     & 21.23 & 16.22 & 26.65 & 18.57 & 14.58 & 11.56 & 30.09 & 26.58 & 23.14 & 18.23 \\
 & w/o Self-Dist   & 21.51 & 16.24 & 23.86 & 17.85 & 10.80 & 8.61  & 29.84 & 25.99 & 21.50 & 17.17 \\
 & w/o Cal \& Dist & 21.05 & 15.51 & 20.02 & 15.65 & 10.36 & 8.79  & 29.42 & 25.44 & 20.21 & 16.35 \\
 & \textbf{MemSuit} & \textbf{22.63} & \textbf{16.36} & \textbf{28.66} & \textbf{20.40} & \textbf{15.40} & \textbf{13.15} & \textbf{33.48} & \textbf{29.96} & \textbf{25.04} & \textbf{19.97} \\
\bottomrule
\end{tabular}%
}
\caption{Ablation study of MemSuit. Removing either component hurts performance.}
\vspace{-10pt}
\label{tab:ablation}
\end{table*}

\noindent\textbf{Dataset.} 
We evaluate on \textbf{LoCoMo}~\cite{maharana2024evaluating}, a benchmark designed to probe long-term conversational dependencies in LLMs. It contains 10 conversations of 200-400 turns each, featuring complex temporal shifts and interleaved topics, paired with 1{,}986 evaluation questions spanning four reasoning categories: multi-hop, temporal, open-domain, and single-hop. We use the first two conversations for training, the third for validation, and the remaining seven for evaluation.

\noindent\textbf{Baselines.}
We compare MemSuit against five representative memory-augmented systems: LoCoMo~\cite{maharana2024evaluating}, MemoryBank~\cite{zhong2024memorybank}, A-Mem~\cite{xu2026mem}, Mem0~\cite{chhikara2025mem0}, and SimpleMem~\cite{liu2026simplemem}.

\noindent\textbf{Backbone Models.}
We instantiate MemSuit with two backbone LLMs: \texttt{Llama-3.1-8B-Inst} \texttt{ruct}~\cite{grattafiori2024llama} and \texttt{Qwen-} \texttt{2.5-3B-Instruct}~\cite{hui2024qwen2}, each serving as both summarizer and reader. The retriever is built on \texttt{all-MiniLM-L6-v2}~\cite{wang2020minilm}.

\noindent\textbf{Implementation Details.}
Conversations are segmented into blocks of 20 turns prior to summarization, and memory entries are indexed in LanceDB. At inference, the retriever returns the top $K{=}10$ entries per query. Both the summarizer and the retriever are fine-tuned for 3 epochs, where we use a learning rate of $2{\times}10^{-5}$ for \texttt{Llama-3.1-8B-Instruct} and $2{\times}10^{-4}$ for \texttt{Qwen-2.5-3B-Instruct} as the summarizer, and $2{\times}10^{-4}$ for the embedding model. All experiments are run on one NVIDIA A100 80GB GPU.

\noindent\textbf{Evaluation Metrics.}
Following prior work on LoCoMo~\cite{liu2026simplemem}, we report \textbf{F1} and \textbf{BLEU} scores per reasoning category, along with the macro-average across categories.

\subsection{Results and Analyses}
\noindent\textbf{Main Results.}
To address \textbf{RQ1}, Table~\ref{tab:main} compares MemSuit against the five baselines on LoCoMo. We make two following observations. \textbf{(1)} Across both backbones, MemSuit attains the best average F1 and BLEU, surpassing the strongest baseline by $4.60$ points in F1 and $3.37$ points in BLEU, demonstrating its effectiveness. \textbf{(2)} The improvement is most pronounced for temporal questions, where MemSuit surpasses the second-best method by $5.54$ F1 points. This aligns with our preliminary analysis in Figure~\ref{fig:vanilla_vs_utility}, where vanilla summaries discard the fine-grained temporal details that utility-aware distillation successfully recovers.

\begin{figure}
    \centering
    \includegraphics[width=\linewidth]{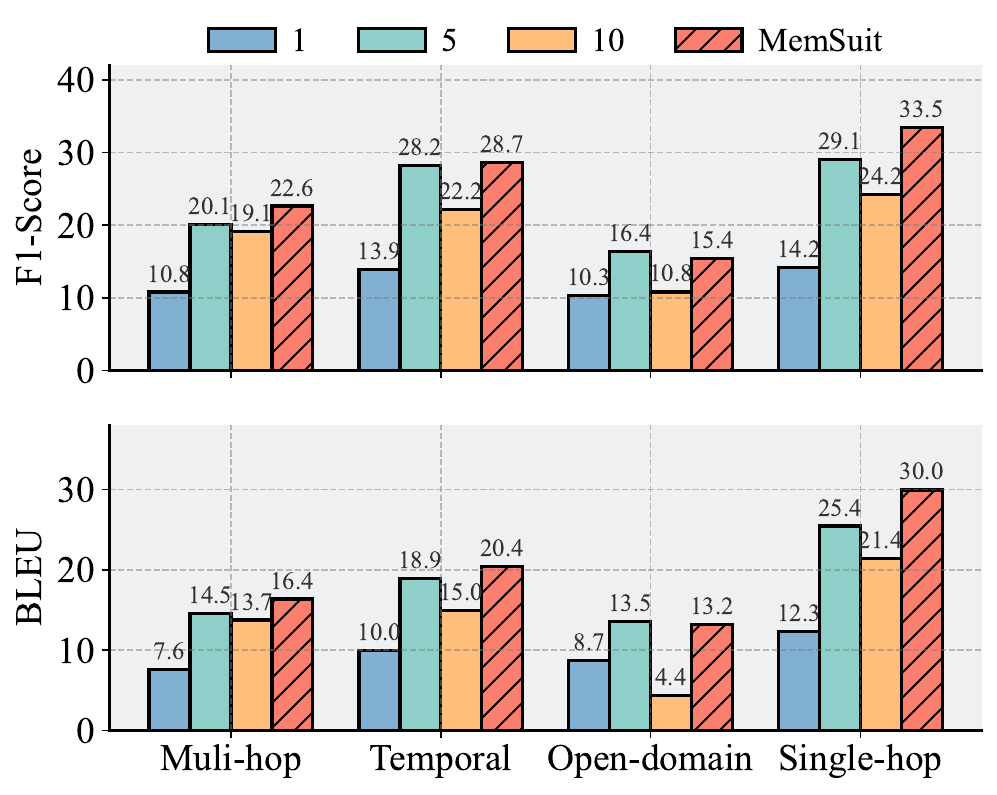}
    \caption{Decomposition granularity ablation on LoCoMo (Qwen2.5-3B). Fixed schemes (1/5/10 turns per entry) underperform MemSuit's adaptive decomposition across all categories of queries.}
    \label{fig:granularity}
\end{figure}

\noindent\textbf{Ablation Study.}
To address the \textbf{RQ2}, We compare MemSuit against three ablated variants: \emph{w/o Rtr-Cal} removes Entry-Aware Retrieval Calibration and uses the off-the-shelf retriever; \emph{w/o Self-Dist} removes Utility-Aware Self-Distillation; and \emph{w/o Cal \& Dist} removes both. From Table~\ref{tab:ablation}, we observe that \textbf{(1)} MemSuit outperforms all variants on macro-average F1 and BLEU under both backbones, confirming that the two components are complementary. \textbf{(2)} Self-distillation contributes more than calibration on temporal and open-domain queries. We attribute this to vanilla summaries omitting the supporting evidence outright, a gap that calibration alone cannot close. 
We further investigate whether the adaptive decomposition is necessary, where we replace the teacher's variable-length output with a fixed scheme that emits exactly one entry per fixed-size block, sweeping block sizes of $1$, $5$, and $10$ turns. As shown in Figure~\ref{fig:granularity}, no fixed setting matches MemSuit, where very fine granularities can fragment evidence and increase retrieval noise, while coarse granularities reintroduce collateral erasure.

\begin{figure}
    \centering
    \includegraphics[width=\linewidth]{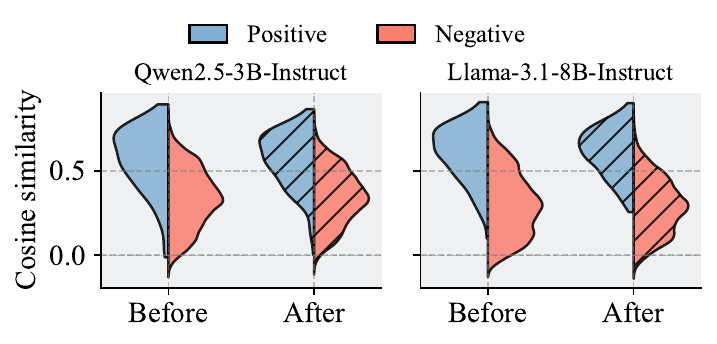}
    \caption{Cosine similarities of positive and negative query-entry pairs before and after calibration. Calibration mainly lifts the tail of the positive distribution.}
    \label{fig:retrieval}
\end{figure}

\noindent\textbf{Retrieval Calibration Analysis.}
To address \textbf{RQ3}, we measure the recall of the entries containing the grounding evidence on the validation set. Table~\ref{tab:retrieval} shows that calibration improves recall at every cutoff, with the largest relative gains concentrated at Recall@1. To localize where this improvement comes from, we further investigate the cosine similarities of positive and negative query-entry pairs before and after calibration in Figure~\ref{fig:retrieval}. We observe that the negative distribution shifts only slightly, while the positive distribution shifts with the largest movement at its tail. In other words, the pairs that benefit most are those whose similarity was initially low. We attribute this to queries such as open-domain ones that share little surface overlap with their grounding entries, which the off-the-shelf encoder fails to embed close together but calibration pulls into alignment.

\begin{table}[t]
\centering
\resizebox{\columnwidth}{!}{%
\begin{tabular}{lcccc}
\toprule
\textbf{Model} &  & \textbf{Recall@1} & \textbf{Recall@5} & \textbf{Recall@10} \\
\midrule
\textbf{Llama} & Before & 25.77 & 54.64 & 64.43 \\
                          & After  & 35.05 & 63.40 & 69.59 \\
\midrule
\textbf{Qwen}             & Before & 21.78 & 44.06 & 51.49 \\
                          & After  & 26.73 & 48.02 & 55.94 \\
\bottomrule
\end{tabular}
}
\caption{Recall of grounding-evidence entries before and after Entry-Aware Retrieval Calibration.}
\label{tab:retrieval}
\end{table}

\noindent\textbf{Generalization Analysis.}
To assess whether MemSuit transfers beyond its training distribution (\textbf{RQ4}), we train on LoCoMo and evaluate on a 50-example subset of LongMemEval without any additional adaptation. We compare against Mem0 and SimpleMem, the strongest baselines from Table~\ref{tab:main}. As shown in Table~\ref{tab:generalization}, MemSuit reaches $28.16$ F1 and $20.84$ BLEU, on par with the best-performing SimpleMem and substantially above Mem0. However, the narrowing of MemSuit's advantage is expected, as LongMemEval's query distribution differs from LoCoMo's, leading the teacher to provide a less precisely targeted distillation signal.

\begin{table}[h]
\centering
\begin{tabular}{lrr}
\toprule
\textbf{Longmemeval} & \textbf{F1} & \textbf{Bleu1} \\
\midrule
Mem0      & 8.92  & 7.01  \\
SimpleMem & \textbf{28.91} & \textbf{21.40}  \\
\textbf{MemSuit}   & 28.16 & 20.84 \\
\bottomrule
\end{tabular}
\caption{Generalization on a 50-example LongMemEval subset with MemSuit trained on LoCoMo. MemSuit matches the strongest baseline despite distribution shift.}
\label{tab:generalization}
\end{table}

\begin{figure}
    \centering
    \includegraphics[width=\linewidth]{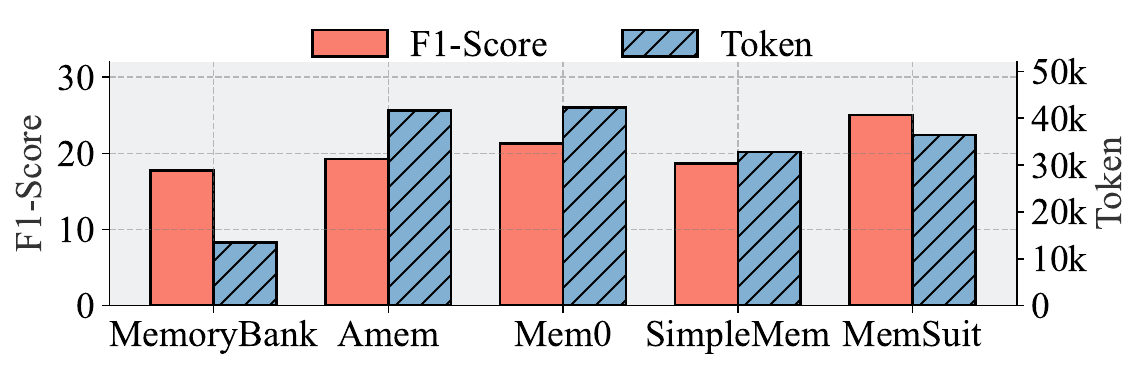}
    \caption{F1 vs.\ summary tokens per conversation on LoCoMo. MemSuit attains the highest F1 at a competitive token budget.}
    \label{fig:token}
\end{figure}

\noindent\textbf{Storage Efficiency Analysis.}
To address \textbf{RQ5}, we measure the average number of summary tokens produced per conversation. Figure~\ref{fig:token} reports token consumption with F1 for each method. We observe that MemSuit produces more compact memory than verbose-summary baselines such as Mem0 while still exceeding them in answer quality, showing that evidence coverage can be improved without a proportional increase in token consumption.

\section{Related Works}

\paragraph{Memory Summarization for LLM Agents.}
Modern memory systems compress past interactions into compact entries that can be retrieved on demand~\cite{wu2025human}. Early \emph{rule-based} methods retain predefined patterns such as named entities, dialogue acts, or temporal markers via handcrafted heuristics~\cite{liu2025integrating,gupta2019abstractive}, which are cheap but brittle. Recent systems delegate memory selection to an LLM and adopt a range of heuristic prompting strategies that vary along two axes. The first is what is written to memory: periodic free-form summaries~\cite{zhong2024memorybank,lu2023memochat}, structured facts governed by explicit lifecycle operations~\cite{chhikara2025mem0}, or minimalist prompted notes~\cite{liu2026simplemem}. The second is how memories are organized: as key-value slots~\cite{modarressi2023ret}, relation-aware graphs~\cite{xu2026mem,li2024graphreader}, or temporal scaffolds~\cite{ong2025towards}. Yet all of their utilized summarizers are optimized, explicitly or implicitly, against human-facing criteria such as faithfulness and coverage~\cite{zhang2024benchmarking}, leaving a gap between the summarization objective and how summaries are consumed by a retrieval-and-answer pipeline. MemSuit closes this gap by grounding summarization in downstream utility through self-distillation.

\paragraph{Self-Distillation with Privileged Information.}
Knowledge distillation~\cite{hinton2015distilling} transfers capabilities from a teacher to a student~\cite{gou2021knowledge,xu2024survey}. Self-distillation, where teacher and student share weights or model family, has been applied to compression~\cite{shen2025codi}, reasoning transfer~\cite{zhang2019your}, and self-improvement via teacher-generated data. A critical variant exposes the teacher to auxiliary signals unavailable at inference, which is a form of learning using privileged information~\cite{vapnik2015learning}. This paradigm has been demonstrated in online action detection with future frames~\cite{zhao2022progressive}, multimodal recognition with privileged modalities~\cite{chen2021learning}, and learning-to-rank with privileged features~\cite{sharmanska2013learning}; and has recently been extended to LLM agents via privileged ground-truth values or reasoning traces~\cite{penaloza2026privileged, snell2022learning}. We are the first to apply privileged-information self-distillation to memory summarization: the teacher observes query-answer pairs to produce utility-aware entries, while the student learns to anticipate plausible queries from the raw conversation alone, thereby internalizing the teacher's selection pattern.

\section{Conclusion}
In this paper, we recast memory summarization for LLM agents as a problem of preserving downstream utility rather than human-facing fidelity. Building on this perspective, we propose MemSuit, a self-distillation framework in which a teacher conditioned on observed query-answer pairs decomposes each conversational block into self-contained entries, a student internalizes this selection criterion without query access at storage time, and a contrastively calibrated retriever bridges the stylistic gap to the resulting compact entries. Empirically, MemSuit yields consistent gains over strong baselines on LoCoMo and exhibits modest robustness under distribution shift, indicating that aligning memory summarizers with their true downstream consumers constitutes a better productive design principle for memory-augmented agents.

\section*{Acknowledgment}
This work was supported in part by the National Science Foundation (NSF) under Grants IIS-2144209, IIS-2223769, BCS-2228534, CMMI-2411248, IIBR-2601942, 2125326, 2228533, and 2402438; by the Office of Naval Research (ONR) under Grant N000142412636; by the Commonwealth Cyber Initiative (CCI) under Grant HV-4Q26-073; and by the Northeastern University iSUPER Impact Engine. 

\section*{Limitations}
\noindent\textbf{Sensitivity to query distribution shift.} As our generalization analysis on LongMemEval shows, MemSuit's advantage narrows when the evaluation query distribution diverges from the training distribution. This is an inherent property of any utility-aware approach: the teacher's selection criteria necessarily reflect the queries it was conditioned on. A promising remedy is to augment the training set with synthetic queries generated by an LLM, broadening the coverage of plausible query types and producing a more diverse distillation signal.

\noindent\textbf{Scale of backbones.} Our experiments use two open-source backbones (\texttt{Llama-3.1-8B-Ins} \texttt{truct} and \texttt{Qwen-2.5-3B-Instruct}), a choice primarily bounded by available computational resources. Applying MemSuit to larger backbones such as \texttt{Llama-3.1-70B-Instruct} or \texttt{Qwen-2.5-32B-Instruct}, which are more representative of production-grade memory systems, would help characterize how the benefits of utility-aware distillation scale with model capacity. However, given the consistent gains observed across both backbones in our experiments, we expect MemSuit to remain effective at larger scales.

\noindent\textbf{Simple training strategy.} We train the summarizer and retriever sequentially against shared teacher supervision rather than jointly. More sophisticated strategies such as joint optimization could in principle allow the two components to co-adapt their representations, and we leave this as a promising direction for future work.

\section*{Use of AI Assistants}
We mainly used AI writing assistants to polish the phrasing and grammar of this manuscript. We take full responsibility for all content in this paper.

\bibliography{custom}

@article{packer2023memgpt,
  title={MemGPT: towards LLMs as operating systems.},
  author={Packer, Charles and Fang, Vivian and Patil, Shishir\_G and Lin, Kevin and Wooders, Sarah and Gonzalez, Joseph\_E},
  year={2023},
  publisher={ArXiv}
}

@article{xu2026mem,
  title={A-mem: Agentic memory for llm agents},
  author={Xu, Wujiang and Liang, Zujie and Mei, Kai and Gao, Hang and Tan, Juntao and Zhang, Yongfeng},
  journal={Advances in Neural Information Processing Systems},
  volume={38},
  pages={17577--17604},
  year={2026}
}

@article{edge2024local,
  title={From local to global: A graph rag approach to query-focused summarization},
  author={Edge, Darren and Trinh, Ha and Cheng, Newman and Bradley, Joshua and Chao, Alex and Mody, Apurva and Truitt, Steven and Metropolitansky, Dasha and Ness, Robert Osazuwa and Larson, Jonathan},
  journal={arXiv preprint arXiv:2404.16130},
  year={2024}
}

@inproceedings{maharana2024evaluating,
  title={Evaluating very long-term conversational memory of llm agents},
  author={Maharana, Adyasha and Lee, Dong-Ho and Tulyakov, Sergey and Bansal, Mohit and Barbieri, Francesco and Fang, Yuwei},
  booktitle={Proceedings of the 62nd Annual Meeting of the Association for Computational Linguistics (Volume 1: Long Papers)},
  pages={13851--13870},
  year={2024}
}

@article{wang2025recursively,
  title={Recursively summarizing enables long-term dialogue memory in large language models},
  author={Wang, Qingyue and Fu, Yanhe and Cao, Yanan and Wang, Shuai and Tian, Zhiliang and Ding, Liang},
  journal={Neurocomputing},
  volume={639},
  pages={130193},
  year={2025},
  publisher={Elsevier}
}

@inproceedings{zhong2024memorybank,
  title={Memorybank: Enhancing large language models with long-term memory},
  author={Zhong, Wanjun and Guo, Lianghong and Gao, Qiqi and Ye, He and Wang, Yanlin},
  booktitle={Proceedings of the AAAI conference on artificial intelligence},
  volume={38},
  number={17},
  pages={19724--19731},
  year={2024}
}

@article{chhikara2025mem0,
  title={Mem0: Building production-ready ai agents with scalable long-term memory},
  author={Chhikara, Prateek and Khant, Dev and Aryan, Saket and Singh, Taranjeet and Yadav, Deshraj},
  journal={arXiv preprint arXiv:2504.19413},
  year={2025}
}

@article{liu2025integrating,
  title={Integrating rule-based NLP and large language models for statin information extraction from clinical notes},
  author={Liu, Siru and McCoy, Allison B and Chen, Qingyu and Wright, Adam},
  journal={International journal of medical informatics},
  pages={106104},
  year={2025},
  publisher={Elsevier}
}

@article{gupta2019abstractive,
  title={Abstractive summarization: An overview of the state of the art},
  author={Gupta, Som and Gupta, Sanjai Kumar},
  journal={Expert Systems with Applications},
  volume={121},
  pages={49--65},
  year={2019},
  publisher={Elsevier}
}

@inproceedings{hu2025hiagent,
  title={Hiagent: Hierarchical working memory management for solving long-horizon agent tasks with large language model},
  author={Hu, Mengkang and Chen, Tianxing and Chen, Qiguang and Mu, Yao and Shao, Wenqi and Luo, Ping},
  booktitle={Proceedings of the 63rd Annual Meeting of the Association for Computational Linguistics (Volume 1: Long Papers)},
  pages={32779--32798},
  year={2025}
}

@article{zhang2025survey,
  title={A survey on the memory mechanism of large language model-based agents},
  author={Zhang, Zeyu and Dai, Quanyu and Bo, Xiaohe and Ma, Chen and Li, Rui and Chen, Xu and Zhu, Jieming and Dong, Zhenhua and Wen, Ji-Rong},
  journal={ACM Transactions on Information Systems},
  volume={43},
  number={6},
  pages={1--47},
  year={2025},
  publisher={ACM New York, NY}
}

@article{liu2026simplemem,
  title={SimpleMem: Efficient Lifelong Memory for LLM Agents},
  author={Liu, Jiaqi and Su, Yaofeng and Xia, Peng and Han, Siwei and Zheng, Zeyu and Xie, Cihang and Ding, Mingyu and Yao, Huaxiu},
  journal={arXiv preprint arXiv:2601.02553},
  year={2026}
}

@article{zhang2024benchmarking,
  title={Benchmarking large language models for news summarization},
  author={Zhang, Tianyi and Ladhak, Faisal and Durmus, Esin and Liang, Percy and McKeown, Kathleen and Hashimoto, Tatsunori B},
  journal={Transactions of the Association for Computational Linguistics},
  volume={12},
  pages={39--57},
  year={2024},
  publisher={MIT Press One Broadway, 12th Floor, Cambridge, Massachusetts 02142, USA~…}
}

@article{yang2026beyond,
  title={Beyond Static Summarization: Proactive Memory Extraction for LLM Agents},
  author={Yang, Chengyuan and Sun, Zequn and Wei, Wei and Hu, Wei},
  journal={arXiv preprint arXiv:2601.04463},
  year={2026}
}

@article{zou2026mem,
  title={ES-Mem: Event Segmentation-Based Memory for Long-Term Dialogue Agents},
  author={Zou, Huhai and Sun, Tianhao and He, Chuanjiang and Tian, Yu and Li, Zhenyang and Jin, Li and Liu, Nayu and Zhong, Jiang and Wei, Kaiwen},
  journal={arXiv preprint arXiv:2601.07582},
  year={2026}
}

@article{li2025survey,
  title={A Survey of Long-Document Retrieval in the PLM and LLM Era},
  author={Li, Minghan and Luo, Miyang and Lv, Tianrui and Zhang, Yishuai and Zhao, Siqi and Nie, Ercong and Zhou, Guodong},
  journal={arXiv preprint arXiv:2509.07759},
  year={2025}
}

@article{hinton2015distilling,
  title={Distilling the knowledge in a neural network},
  author={Hinton, Geoffrey and Vinyals, Oriol and Dean, Jeff},
  journal={arXiv preprint arXiv:1503.02531},
  year={2015}
}

@article{gou2021knowledge,
  title={Knowledge distillation: A survey},
  author={Gou, Jianping and Yu, Baosheng and Maybank, Stephen J and Tao, Dacheng},
  journal={International journal of computer vision},
  volume={129},
  number={6},
  pages={1789--1819},
  year={2021},
  publisher={Springer}
}

@article{xu2024survey,
  title={A survey on knowledge distillation of large language models},
  author={Xu, Xiaohan and Li, Ming and Tao, Chongyang and Shen, Tao and Cheng, Reynold and Li, Jinyang and Xu, Can and Tao, Dacheng and Zhou, Tianyi},
  journal={arXiv preprint arXiv:2402.13116},
  year={2024}
}

@inproceedings{shen2025codi,
  title={Codi: Compressing chain-of-thought into continuous space via self-distillation},
  author={Shen, Zhenyi and Yan, Hanqi and Zhang, Linhai and Hu, Zhanghao and Du, Yali and He, Yulan},
  booktitle={Proceedings of the 2025 Conference on Empirical Methods in Natural Language Processing},
  pages={677--693},
  year={2025}
}

@inproceedings{zhang2019your,
  title={Be your own teacher: Improve the performance of convolutional neural networks via self distillation},
  author={Zhang, Linfeng and Song, Jiebo and Gao, Anni and Chen, Jingwei and Bao, Chenglong and Ma, Kaisheng},
  booktitle={Proceedings of the IEEE/CVF international conference on computer vision},
  pages={3713--3722},
  year={2019}
}

@article{vapnik2015learning,
  title={Learning using privileged information: similarity control and knowledge transfer},
  author={Vapnik, Vladimir and Izmailov, Rauf},
  journal={The Journal of Machine Learning Research},
  volume={16},
  number={1},
  pages={2023--2049},
  year={2015},
  publisher={JMLR. org}
}

@article{zhao2022progressive,
  title={Progressive privileged knowledge distillation for online action detection},
  author={Zhao, Peisen and Xie, Lingxi and Wang, Jiajie and Zhang, Ya and Tian, Qi},
  journal={Pattern Recognition},
  volume={129},
  pages={108741},
  year={2022},
  publisher={Elsevier}
}

@article{chen2021learning,
  title={Learning with privileged multimodal knowledge for unimodal segmentation},
  author={Chen, Cheng and Dou, Qi and Jin, Yueming and Liu, Quande and Heng, Pheng Ann},
  journal={IEEE transactions on medical imaging},
  volume={41},
  number={3},
  pages={621--632},
  year={2021},
  publisher={IEEE}
}

@inproceedings{sharmanska2013learning,
  title={Learning to rank using privileged information},
  author={Sharmanska, Viktoriia and Quadrianto, Novi and Lampert, Christoph H},
  booktitle={Proceedings of the IEEE international conference on computer vision},
  pages={825--832},
  year={2013}
}

@article{penaloza2026privileged,
  title={Privileged Information Distillation for Language Models},
  author={Penaloza, Emiliano and Vattikonda, Dheeraj and Gontier, Nicolas and Lacoste, Alexandre and Charlin, Laurent and Caccia, Massimo},
  journal={arXiv preprint arXiv:2602.04942},
  year={2026}
}

@article{snell2022learning,
  title={Learning by distilling context},
  author={Snell, Charlie and Klein, Dan and Zhong, Ruiqi},
  journal={arXiv preprint arXiv:2209.15189},
  year={2022}
}

@article{lu2023memochat,
  title={Memochat: Tuning llms to use memos for consistent long-range open-domain conversation},
  author={Lu, Junru and An, Siyu and Lin, Mingbao and Pergola, Gabriele and He, Yulan and Yin, Di and Sun, Xing and Wu, Yunsheng},
  journal={arXiv preprint arXiv:2308.08239},
  year={2023}
}

@article{modarressi2023ret,
  title={Ret-llm: Towards a general read-write memory for large language models},
  author={Modarressi, Ali and Imani, Ayyoob and Fayyaz, Mohsen and Sch{\"u}tze, Hinrich},
  journal={arXiv preprint arXiv:2305.14322},
  year={2023}
}

@inproceedings{li2024graphreader,
  title={Graphreader: Building graph-based agent to enhance long-context abilities of large language models},
  author={Li, Shilong and He, Yancheng and Guo, Hangyu and Bu, Xingyuan and Bai, Ge and Liu, Jie and Liu, Jiaheng and Qu, Xingwei and Li, Yangguang and Ouyang, Wanli and others},
  booktitle={Findings of the Association for Computational Linguistics: EMNLP 2024},
  pages={12758--12786},
  year={2024}
}

@inproceedings{ong2025towards,
  title={Towards lifelong dialogue agents via timeline-based memory management},
  author={Ong, Kai Tzu-iunn and Kim, Namyoung and Gwak, Minju and Chae, Hyungjoo and Kwon, Taeyoon and Jo, Yohan and Hwang, Seung-won and Lee, Dongha and Yeo, Jinyoung},
  booktitle={Proceedings of the 2025 Conference of the Nations of the Americas Chapter of the Association for Computational Linguistics: Human Language Technologies (Volume 1: Long Papers)},
  pages={8631--8661},
  year={2025}
}

@article{wu2025human,
  title={From human memory to ai memory: A survey on memory mechanisms in the era of llms},
  author={Wu, Yaxiong and Liang, Sheng and Zhang, Chen and Wang, Yichao and Zhang, Yongyue and Guo, Huifeng and Tang, Ruiming and Liu, Yong},
  journal={arXiv preprint arXiv:2504.15965},
  year={2025}
}

@article{grattafiori2024llama,
  title={The llama 3 herd of models},
  author={Grattafiori, Aaron and Dubey, Abhimanyu and Jauhri, Abhinav and Pandey, Abhinav and Kadian, Abhishek and Al-Dahle, Ahmad and Letman, Aiesha and Mathur, Akhil and Schelten, Alan and Vaughan, Alex and others},
  journal={arXiv preprint arXiv:2407.21783},
  year={2024}
}

@article{hui2024qwen2,
  title={Qwen2. 5-coder technical report},
  author={Hui, Binyuan and Yang, Jian and Cui, Zeyu and Yang, Jiaxi and Liu, Dayiheng and Zhang, Lei and Liu, Tianyu and Zhang, Jiajun and Yu, Bowen and Lu, Keming and others},
  journal={arXiv preprint arXiv:2409.12186},
  year={2024}
}

@article{wang2020minilm,
  title={Minilm: Deep self-attention distillation for task-agnostic compression of pre-trained transformers},
  author={Wang, Wenhui and Wei, Furu and Dong, Li and Bao, Hangbo and Yang, Nan and Zhou, Ming},
  journal={Advances in neural information processing systems},
  volume={33},
  pages={5776--5788},
  year={2020}
}

\newpage

\appendix
%
%
%
%
%
%

\appendix
%
%
%

\appendix

\section{Additional Experiments}
\label{app:rebuttal}

\subsection{Judge Model Sensitivity}
\label{app:rebuttal:judge}

To assess whether MemSuit's performance depends on the specific model used as the grounding-entry judge $\psi$, we swap the judge from \texttt{Qwen2.5-3B-Instruct} to \texttt{Llama-3.1-8B-Inst} \texttt{ruct} while keeping the Qwen backbone fixed for the summarizer and reader. Table~\ref{tab:judge_sensitivity} shows that performance remains stable across judges, confirming that MemSuit is robust to this choice.

\begin{table}[h]
\centering
\resizebox{\columnwidth}{!}{%
\begin{tabular}{lcc}
\toprule
\textbf{Metric} & \textbf{Qwen2.5-3B} & \textbf{Llama-3.1-8B} \\
\midrule
Multi-F1    & 22.63 & 22.89 \\
Multi-BLEU  & 16.36 & 15.22 \\
Temp-F1     & 28.66 & 27.78 \\
Temp-BLEU   & 20.40 & 18.39 \\
Open-F1     & 15.40 & 17.15 \\
Open-BLEU   & 13.15 & 13.96 \\
Single-F1   & 33.48 & 32.24 \\
Single-BLEU & 29.96 & 28.68 \\
\bottomrule
\end{tabular}%
}
\caption{Sensitivity of MemSuit (Qwen2.5-3B backbone) to the choice of grounding-entry judge $\psi$. Results are stable across judge models.}
\label{tab:judge_sensitivity}
\end{table}

\subsection{Isolating Decomposition from Query Conditioning}
\label{app:rebuttal:nontarget}

To disentangle the effect of adaptive entry decomposition from that of query conditioning, we construct a non-target variant in which the teacher performs the same adaptive decomposition but is not conditioned on the query--answer pair. Table~\ref{tab:nontarget} compares this variant against full MemSuit (Qwen2.5-3B backbone). The non-target variant falls consistently behind full MemSuit, confirming that decomposition alone does not account for the gains and that the query-conditioned, utility-aware signal is essential.

\begin{table}[h]
\centering
\resizebox{\columnwidth}{!}{%
\begin{tabular}{lcc}
\toprule
\textbf{Metric} & \textbf{MemSuit} & \textbf{Adaptive (non-target)} \\
\midrule
Multi-F1    & 22.63 & 19.62 \\
Multi-BLEU  & 16.36 & 13.82 \\
Temp-F1     & 28.66 & 24.75 \\
Temp-BLEU   & 20.40 & 19.57 \\
Open-F1     & 15.40 & 15.85 \\
Open-BLEU   & 13.15 & 12.10 \\
Single-F1   & 33.48 & 31.88 \\
Single-BLEU & 29.96 & 28.43 \\
\bottomrule
\end{tabular}%
}
\caption{Ablation isolating adaptive decomposition from query conditioning (Qwen2.5-3B backbone). The non-target variant performs adaptive decomposition without conditioning on the query--answer pair.}
\label{tab:nontarget}
\end{table}

\subsection{In-Domain Training on LongMemEval}
\label{app:rebuttal:longmemeval_id}

Our main generalization analysis (Table~\ref{tab:generalization}) trains MemSuit on LoCoMo and evaluates it on LongMemEval without adaptation. To further test whether MemSuit's effectiveness is specific to LoCoMo, we additionally train MemSuit directly on LongMemEval (first 50--150 examples) and evaluate on the same 50-example subset used in Table~\ref{tab:generalization}. As shown in Table~\ref{tab:longmemeval_id}, once trained in-domain, MemSuit again surpasses the strongest baseline, confirming that its gains generalize beyond a single training dataset.

\begin{table}[h]
\centering
\resizebox{\columnwidth}{!}{%
\begin{tabular}{lrr}
\toprule
\textbf{LongMemEval} & \textbf{F1} & \textbf{BLEU-1} \\
\midrule
Mem0                                  & 8.92  & 7.01  \\
SimpleMem                             & 28.91 & 21.40 \\
MemSuit (LoCoMo-trained, OOD)         & 28.16 & 20.84 \\
\textbf{MemSuit (LongMemEval-trained, ID)} & \textbf{30.89} & \textbf{24.32} \\
\bottomrule
\end{tabular}%
}
\caption{In-domain training on LongMemEval. MemSuit trained directly on LongMemEval surpasses both the out-of-domain (LoCoMo-trained) variant and the strongest baseline.}
\label{tab:longmemeval_id}
\end{table}

\subsection{Generalization to New Backbones}
\label{app:rebuttal:newer_backbone}

To evaluate whether MemSuit's gains carry over to newer model architectures, we instantiate MemSuit and all baselines with \texttt{Qwen3-1.7B} on LoCoMo. Table~\ref{tab:qwen3} shows that MemSuit again achieves the best average performance, confirming that its advantages are not specific to the two-year-old backbones used in the main experiments.

\begin{table*}[t]
\centering
\renewcommand{\arraystretch}{1.15}
\setlength{\tabcolsep}{4pt}
\resizebox{\textwidth}{!}{%
\begin{tabular}{l cccccc}
\toprule
\textbf{Metric} & \textbf{LoCoMo} & \textbf{MemoryBank} & \textbf{A-mem} & \textbf{Mem0} & \textbf{SimpleMem} & \textbf{MemSuit} \\
\midrule
Multi-F1    & 8.51 & 9.91  & 18.22 & 15.44 & 17.56 & \textbf{19.88} \\
Multi-BLEU  & 6.96 & 7.21  & 14.91 & 11.27 & 12.77 & \textbf{15.22} \\
Temp-F1     & 6.34 & 12.07 & 23.86 & 16.08 & 22.39 & \textbf{23.87} \\
Temp-BLEU   & 5.14 & 9.04  & \textbf{18.60} & 12.20 & 15.15 & 15.55 \\
Open-F1     & 9.48 & 6.97  & 15.09 & 15.35 & 16.27 & \textbf{22.88} \\
Open-BLEU   & 7.36 & 5.24  & 12.83 & 12.25 & 12.70 & \textbf{19.52} \\
Single-F1   & 9.82 & 13.10 & 27.79 & 26.17 & 27.56 & \textbf{28.50} \\
Single-BLEU & 8.69 & 8.62  & 21.72 & 21.62 & 21.35 & \textbf{24.85} \\
\midrule
Avg-F1      & 8.54 & 10.51 & 21.24 & 18.26 & 20.95 & \textbf{23.78} \\
Avg-BLEU    & 7.04 & 7.53  & 17.02 & 14.34 & 15.49 & \textbf{18.79} \\
\bottomrule
\end{tabular}%
}
\caption{Results on LoCoMo with a newer-generation backbone, \texttt{Qwen3-1.7B}, for both MemSuit and all baselines. MemSuit maintains the best average F1 and BLEU.}
\label{tab:qwen3}
\end{table*}

\section{Experimental Details}
\label{app:exp_details}

\subsection{Dataset and Splits}
\label{app:dataset}

We partition the ten LoCoMo conversations at the conversation level to ensure that no block, speaker pair, or topical thread leaks across splits. Conversations \texttt{0}--\texttt{1} are used for training, conversation \texttt{2} for validation, and conversations \texttt{3}--\texttt{9} for testing. The evidence-coverage analysis in Figure~\ref{fig:vanilla_vs_utility} is conducted on conversations \texttt{2} and \texttt{4}.

\subsection{Rule-Based Summarizer Baseline}
\label{app:rule_based}

The rule-based baseline shown in Figure~\ref{fig:teaser}(a) and Figure~\ref{fig:vanilla_vs_utility} performs two extraction passes per turn. The first pass extracts atomic facts via regular expressions for dates, times, numerals, and relative-time expressions, complemented by named-entity recognition; each match yields a single entry. The second pass extracts subject-verb-object triples through spaCy dependency parsing, producing one proposition-level entry per triple. The triple pass elevates this baseline beyond a bag-of-tokens extractor: rather than emitting isolated entity entries such as \texttt{[ENTITY] Maria} and \texttt{[ENTITY] car}, it emits bound propositions such as \texttt{(Maria, donate, old car)}. When spaCy or its English model is unavailable, the system falls back to closed-vocabulary regular expressions matching simple \texttt{Subject-Verb-Object} templates.

\subsection{Evidence Coverage Protocol}
\label{app:evidence_coverage}

Evidence coverage measures, for each query, whether the memory entries produced by a summarizer contain sufficient information to support the gold answer. For each query, we run the summarizer over its source conversation, collect the resulting entries, and prompt \texttt{Qwen2.5-3B-Instruct} as an LLM judge to issue a binary YES/NO verdict on whether the entries jointly support the gold answer. The judge is explicitly instructed to allow surface-form variation (e.g., date normalization), multi-hop composition across entries, and reasonable inference, while rejecting cases requiring external world knowledge. Coverage is reported per category as the fraction of queries receiving a YES verdict. The full judge prompt is given below.

\begin{promptbox}{Evidence Coverage Judge Prompt}
\begin{Verbatim}[fontsize=\scriptsize, breaklines=true, breakanywhere=true]
You are given a question, its gold answer, and a list
of memory notes extracted from a conversation. Decide
whether the notes contain enough information to
support the gold answer.

Memory notes:
{formatted_entries}

Question:    {question}
Gold answer: {gold_answer}

Decision rules:
- Answer YES if the notes - taken together, with
  reasonable inference and composition - establish the
  gold answer.
- Answer YES even when surface forms differ. For
  example:
    * "May 7, 2023", "7 May 2023", "2023-05-07" all
      match the same date.
    * "$680" matches "680 dollars" or "680 USD".
    * "Caroline Smith" matches "Caroline" when context
      makes it unambiguous.
- Answer YES if the gold answer can be obtained by
  composing several notes (multi-hop reasoning over
  the notes).
- Answer YES if the gold answer is a reasonable
  inference grounded in the notes (e.g., the question
  asks what someone "might" do, and the notes describe
  their relevant interests, skills, or stated plans).
- Answer NO only if the specific fact required by the
  gold answer is missing from the notes, or if
  establishing the gold answer would require
  information not present in the notes.
- For gold answers with multiple components (e.g.,
  "A, B"), answer YES if the notes support each
  component.
- Do NOT use general world knowledge that is unrelated
  to the conversation.

First, write 1-3 short sentences of reasoning citing
the relevant note numbers. Then end your reply with
exactly one of these two lines:
Verdict: YES
Verdict: NO
\end{Verbatim}
\end{promptbox}

\subsection{Code Availability}
\label{app:code}

Our official code for MemSuit is available at \href{https://github.com/LzyFischer/MemSuit}{https://github.com/LzyFischer/MemSuit}.

\section{Training and Inference Configuration}
\label{app:training}

\paragraph{Self-Distillation.}
We fine-tune the student summarizer with LoRA (rank $r{=}16$, $\alpha{=}32$, dropout $0.05$), a per-device batch size of $2$ with gradient accumulation of $8$ (effective batch size $16$), and \texttt{bf16} mixed precision. We use a warmup ratio of $0.03$ followed by linear decay over $3$ epochs. The learning rate is $2{\times}10^{-5}$ for \texttt{Llama-3.1-8B-Instruct} and $2{\times}10^{-4}$ for \texttt{Qwen2.5-3B-Instruct}.

\paragraph{Retrieval Calibration.}
We fine-tune \texttt{all-Min} \texttt{iLM-L6-v2} with a batch size of $32$ under \texttt{bf16} mixed precision, a learning rate of $2{\times}10^{-4}$, and the same $0.03$ warmup ratio for $3$ epochs.

\paragraph{Inference.}
All inference, including teacher generation, student generation, and reader QA, is served through vLLM. The retriever returns the top $K{=}10$ entries per query, and all memory entries are indexed in LanceDB.

\section{Prompts}
\label{app:prompts}

This section provides the complete set of prompts used by MemSuit's pipeline components. We group them by lifecycle stage: \emph{training-time} prompts that produce teacher and judge supervision (Appendix~\ref{app:prompts:train}), and \emph{inference-time} prompts for the deployed student summarizer and the fixed reader (Appendix~\ref{app:prompts:infer}). The evaluation-time prompt for evidence-coverage judging is co-located with its protocol in Appendix~\ref{app:evidence_coverage}.

\subsection{Training-Time Prompts}
\label{app:prompts:train}

\subsubsection{Teacher: Utility-Aware Decomposition}
\label{app:prompts:teacher}

The teacher receives the dialogue block together with an observed query--answer pair. The prompt is deliberately structured so that the query--answer pair acts purely as a post-hoc coverage check, applied after a comprehensive set of entries has already been drafted. This design discourages the teacher from collapsing the output around the conditioning query, mitigating the collateral-erasure problem identified in Section~\ref{subsec:entry-split}.

\begin{promptbox}{Teacher Prompt --- Utility-Aware Decomposition}
\begin{Verbatim}[fontsize=\scriptsize, breaklines=true, breakanywhere=true]
[Internal Coverage Check - do not mention in output,
do not let it shape your focus]
A downstream system will later be asked the following
question against your memory entries. This block exists
ONLY to verify coverage after you have already drafted
a comprehensive set of entries. It is NOT the topic of
this extraction and must NOT cause you to drop,
shorten, merge, reorder, or de-prioritize any other
entry.

  Question:    {question}
  Gold answer: {answer}
  Evidence utterance (must be preserved verbatim in
  at least one entry):
    {evidence_text}

Procedure:
1. First, draft your memory entries normally, covering
   ALL information in the dialogue at the level of
   detail required by the main Requirements above
   (about one entry per dialogue turn). Do this as if
   this block did not exist.
2. Then, verify that the evidence utterance above
   appears, with its original wording preserved (key
   nouns, verbs, and named entities unchanged), inside
   the lossless_restatement of at least one entry -
   and that the gold answer is recoverable from that
   entry using only the original words from the
   dialogue.
3. If the check already passes, change nothing. If it
   does not, add or minimally revise exactly ONE entry
   to satisfy it, and leave every other entry
   untouched.

Hard constraints:
- Do NOT mention the question, the gold answer, the
  evidence, or this check in your output.
- Do NOT phrase any entry as an answer to the question;
  entries are statements about the dialogue, not
  responses.
- Do NOT use the gold answer's wording if it does not
  appear in the dialogue - use the original word(s)
  from the raw text.
- Coverage of unrelated topics (small talk, emotions,
  plans, image descriptions, etc.) must remain
  identical to what you would produce without this
  block. Fewer entries than the dialogue's richness
  warrants is a failure of this task.
\end{Verbatim}
\end{promptbox}

\subsubsection{LLM Judge: Grounding-Entry Identification}
\label{app:prompts:judge}

Given the teacher's decomposed entries for a block, the judge $\psi$ selects the single entry that most directly supports the gold answer. The selected entry is used as the positive example in the contrastive objective of Section~\ref{subsec:contrastive}.

\begin{promptbox}{LLM Judge Prompt --- Grounding-Entry Identification}
\begin{Verbatim}[fontsize=\scriptsize, breaklines=true, breakanywhere=true]
We are building a retrieval-augmented memory system.
At inference time, the system will see only the
question and must retrieve the most relevant memory
entry from the store. We need to label which entry is
the correct retrieval target.

[Question]
{question}

[Gold Answer]
{answer}

[Evidence Turn - the dialogue line this question is
grounded in]
{evidence_text}

[Candidate Memory Entries]
{numbered_entries}

[Your Task]
Pick the index (0-based) of the SINGLE entry that:
  (a) directly contains the fact that grounds the gold
      answer, AND
  (b) would be the most useful entry to retrieve given
      ONLY the question (without seeing the evidence
      or answer).

If two entries both ground the answer, prefer the one
whose phrasing is more self-contained and more likely
to match a query that asks the question.

[Output Format]
Return ONLY a JSON object of the form:
  {"index": <int>}
where <int> is in the range [0, {n_entries_minus_1}].
No prose, no markdown.
\end{Verbatim}
\end{promptbox}

\subsection{Inference-Time Prompts}
\label{app:prompts:infer}

\subsubsection{Student: Memory Entry Generation}
\label{app:prompts:student}

At inference time, the student receives the dialogue block alone, without any query conditioning, and must emit a self-contained set of memory entries in JSON form.

\begin{promptbox}{Student Prompt --- Memory Entry Generation}
\begin{Verbatim}[fontsize=\scriptsize, breaklines=true, breakanywhere=true]
Your task is to extract all valuable information from
the following dialogues and convert them into
structured memory entries.

[Current Window Dialogues]
{dialogue_text}

[Requirements]
1. Complete Coverage: Generate enough memory entries
   to ensure ALL information in the dialogues is
   captured.
2. Force Disambiguation: Absolutely PROHIBIT using
   pronouns (he, she, it, they, this, that) and
   relative time (yesterday, today, last week,
   tomorrow). Use full names and absolute ISO 8601
   timestamps inline.
3. Lossless Information: Each entry's
   lossless_restatement must be a complete,
   independent, understandable sentence that includes
   all relevant subjects, objects, time, and location
   inline.

[Output Format]
Return a JSON array. Each element is a memory entry
with a single field:

[
  {
    "lossless_restatement": "Complete unambiguous
    restatement (must include all subjects, objects,
    time, location, etc.)"
  },
  ...
]

Now process the above dialogues. Return ONLY the JSON
array, no other explanations.
\end{Verbatim}
\end{promptbox}

\subsubsection{Reader: Question Answering}
\label{app:prompts:reader}

The fixed reader $\pi$ receives the query and the top-$K$ retrieved entries, and produces a concise answer together with a brief reasoning chain in JSON form.

\begin{promptbox}{Reader Prompt --- Question Answering}
\begin{Verbatim}[fontsize=\scriptsize, breaklines=true, breakanywhere=true]
Answer the user's question based on the provided
context.

User Question: {query}

Relevant Context:
{context_str}

Requirements:
1. First, think through the reasoning process.
2. Then provide a very CONCISE answer (short phrase
   about core information).
3. Answer must be based ONLY on the provided context.
4. All dates in the response must be formatted as
   'DD Month YYYY' but you can output more or less
   details if needed.
5. Return your response in JSON format.

Output Format:
{
  "reasoning": "Brief explanation of your thought
                process",
  "answer":    "Concise answer in a short phrase"
}

Now answer the question. Return ONLY the JSON, no
other text.
\end{Verbatim}
\end{promptbox}

\end{document}